# Enhancing Image Retrieval: A Comprehensive Study on Photo Search using the CLIP Mode

Harini S
Assistant Professor
Dept. of Information Science and Engineering,
BMS College of Engineering
Bengaluru, India
harini.ise@bmsce.ac.in

Naresh Kumar Lahajal
Dept. of Information Science and Engineering,
BMS College of Engineering
Bengaluru, India
naresh.is21@bmsce.ac.in

*Abstract* - Photo search, the task of retrieving images based on textual queries, has witnessed significant advancements with the introduction of CLIP (Contrastive Language-Image Pretraining) model. CLIP leverages a vision-language pre training approach, wherein it learns a shared representation space for images and text, enabling cross-modal understanding. This model demonstrates the capability to understand the semantic relationships between diverse image and text pairs, allowing for efficient and accurate retrieval of images based on natural language queries. By training on a large-scale dataset containing images and their associated textual descriptions, CLIP achieves remarkable generalization, providing a powerful tool for tasks such as zero-shot learning and few-shot classification. This abstract summarizes the foundational principles of CLIP and highlights its potential impact on advancing the field of photo search, fostering a seamless integration of natural language understanding and computer vision for improved information retrieval in multimedia applications.

## INTRODUCTION

Photo search using the CLIP (Contrastive Language-Image Pretraining) model represents a paradigm shift in the way computers understand and interpret images in conjunction with natural language. The fundamental idea behind CLIP is the development of a unified representation space where both images and textual descriptions coexist, allowing for a seamless interaction between the two modalities.

In a traditional photo search scenario, the challenge lies in bridging the gap between the ways humans express their information needs through language and the visual content stored in images. CLIP addresses this challenge by training on a vast dataset that includes diverse pairs of images and associated textual descriptions. During this training, the model learns to map images and text into a common feature space, where semantically related image-text pairs are brought closer together.

The power of CLIP lies in its ability to generalize its understanding across a wide range of concepts, even those not explicitly encountered during training. This makes it particularly effective for zero-shot learning, where the model can perform tasks it has never seen before by leveraging its understanding of the relationships between images and text. Moreover, the few-shot learning capability allows CLIP to adapt quickly to new tasks with minimal examples, making it a versatile tool for various applications.

In the context of photo search, users can input natural language queries, and CLIP retrieves images that semantically match the query. The model's understanding extends beyond mere visual similarity, encompassing nuanced relationships and contextual meanings conveyed through language. This makes CLIP a robust solution for tasks that demand a deeper understanding of the content within images, facilitating more intuitive and accurate photo search experiences. The unified vision-language representation learned by CLIP has broader implications, fostering advancements in multimedia applications, content retrieval, and cross-modal understanding in artificial intelligence

Research Questions and Objectives:

How does the CLIP model enhance semantic understanding in the context of image retrieval?

What is the extent of CLIP's proficiency in zero-shot learning scenarios, and how does it generalize to unseen concepts during image retrieval?

What are the strengths and limitations of CLIP in comparison to existing image retrieval methods?

How can the findings of this research contribute to refining and maximizing the benefits of CLIP in the broader landscape of photo search and digital content retrieval.?

## LITERATURE REVIEW

In the dynamic landscape of image retrieval, the CLIP model introduced by Dosovitskiy et al. has emerged as a game-changer, seamlessly intertwining text and images for classification [1]. This visionary approach has set the stage for a paradigm shift in image understanding. Building upon this foundation, the NeurIPS 2020 paper on learning transferable visual models emphasizes the intricate dynamics involved in training large-scale datasets, shedding light on the interplay between diverse language descriptions and visual representations [2]. This not only refines our understanding of the training process but also underscores the importance of harnessing textual information to enhance visual comprehension.

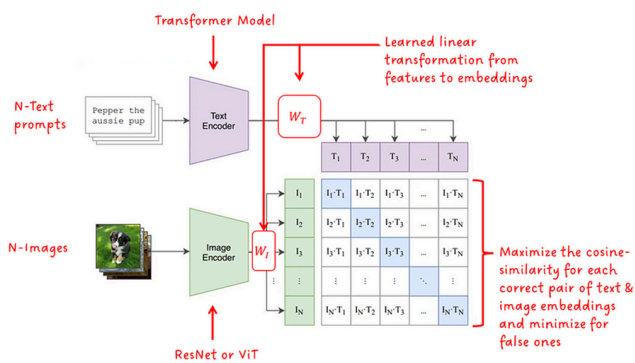

Fig. 1 — CLIP's Architecture and training process. Image Source + annotations by Sascha Kirch

The model architecture consists of two encoder models, one for each modality. For the text encoder a transformer was used while the image encoder uses either a version of ResNet or ViT (Vision Transformer). A learned linear transformation, one for each modality, transforms the features into embeddings of matching size. Finally, the cosine similarity is calculated between each of the embeddings of opposing modality and is scaled by a learned temperature scalar. During training, the cosine similarity between matching pairs is maximized while it is minimized for incorrect pairs, hence the term "contrastive" in the framework's name.

There are some subtleties that are crucial for the success, beside the large dataset of course. First, the contrastive learning approach strongly depends on the batch size N. The more negative samples are provided along the correct ones, the stronger the learning signal. CLIP was trained on a batch size of 32,768, which is quite large. Second, CLIP does not learn a match of the exact wording, but an easier proxy task to only learn the text as a whole, also called bag of words (BoW).

***Fun Fact:*** *The version of CLIP using a ResNet50x64 as image encoder was trained for 18 days on 592 V100 GPUS and while the version with the ViT model was trained for 12 days on 256 V100 GPUS. In other words,* **over 29 years** *and* **over 8 years** *on a single GPU respectively (ignoring the fact a different batch size would be used).*

Once the model is trained it can be used to perform object classification on images. The question is: how to perform classification using a model that has not been trained to classify images nor does input class labels but text prompts? Fig 2. shows how:

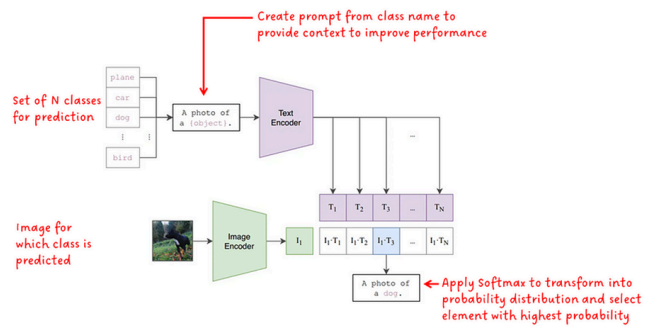

Fig. 2 — CLIP's Architecture for image classification. Image Source + annotations by Sascha Kirch

A class label can be seen as a text prompt formed by a single word. To tell the model which classes are available for the classification task, a set of N classes is input into the model. This is a huge advantage compared to classification models trained on a fixed set of labels. We can now either input 3 classes or 100; it's our choice. As we will see later, to improve the performance of CLIP, the class label is transformed into a prompt to provide further context to the model. Each prompt is then fed to the text encoder and is then transformed into an embedding vector. The input image is fed into the image encoder to obtain the embedding vector.

Then the cosine similarity is calculated for each pair of text and image embeddings. A SoftMax is applied on the obtained similarity values to form a probability distribution. Finally, the value with the highest probability is selected as the final prediction.

The CLIP paper presents a vast number of experiments and ablations. Here we will cover five, from which I think are important to understand the success of CLIP. Upfront the

take aways (as formulated by the authors of CLIP) and then we will dive into the details:

1. **Training Efficiency:** CLIP is much more efficient at zero-shot transfer than our image caption baseline
2. **Text Input Format:** Prompt engineering and ensembling improve zero-shot performance
3. **Zero-Shot Performance:** Zero-shot CLIP is competitive with fully super-vised baseline
4. **Few-Shot Performance:** Zero-shot CLIP outperforms few-shot linear probes
5. **Distribution Shift:** Zero-shot CLIP is much more robust to distribution shift than standard ImageNet models

**Training Efficiency**

During training, the image encoder and the text encoder are trained jointly, meaning with a single training objective and at the same time. Not only does CLIP perform a contrastive learning scheme, but the text prompts are compared as a whole against a given image, hence the order of words does not matter. It is simply a "bag of words". The phrase "my name is Sascha" results in the same embedding as "Sascha name is my".

Predicting a bag of words instead of the correct words and its position in a phrase is a much easier proxy objective. Fig 3. below shows the zero-shot accuracy on ImageNet over the number of training samples of the initial transformer model trained to predict exact words, the initial transformer model trained to predict a bag of words and the CLIP model that performs contrastive learning using bag of words.

*"CLIP is much more efficient at zero-shot transfer than our image caption baseline" — CLIP Authors*

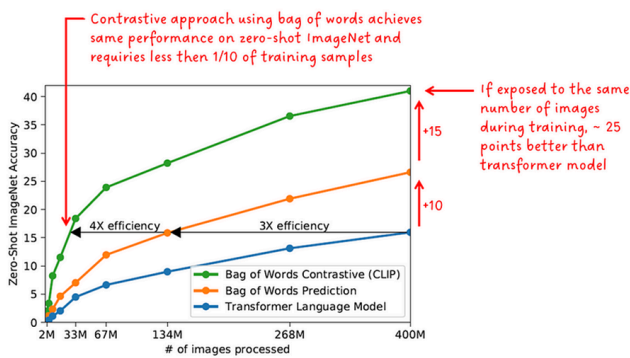

Fig. 3 — Zero-shot efficiency. Image Source + annotations by Sascha Kirch

**Text Input Format**

As we have seen in Fig. 2, to perform object classification, the class label has been converted into a text prompt. Of course, this was not by chance, because CLIP would be totally fine with a single word. It was done to leverage the descriptiveness of language and to provide context to resolve possible ambiguities. Let's take the word "boxer" for example. It could be a type of dog or a type of athlete. The authors of CLIP have shown that the format of the text prompt matters a lot and can boost the performance as well increase the efficiency.

*"Prompt engineering and ensembling improve zero-shot performance" — CLIP Authors*

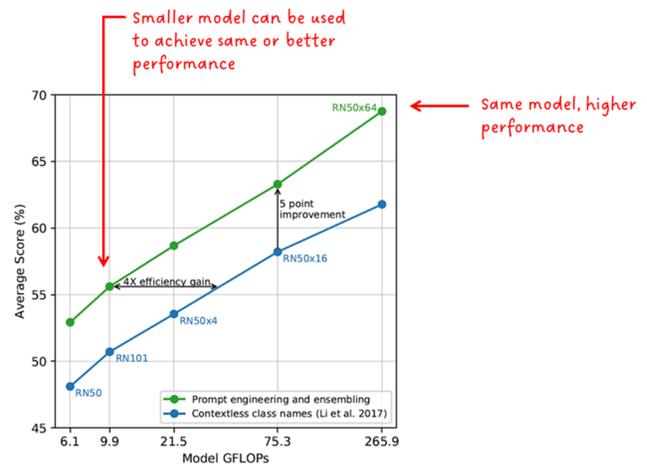

Fig. 4— Prompt engineering and ensembling vs. contextless class names. Image Source + annotations by Sascha Kirch

Zero-Shot Performance

In another experiment, the authors compared the zero-shot image classification performance of CLIP against a model that was trained specifically on the dataset under comparison.

*"Zero-shot CLIP is competitive with fully super-vised baseline" — CLIP Authors*

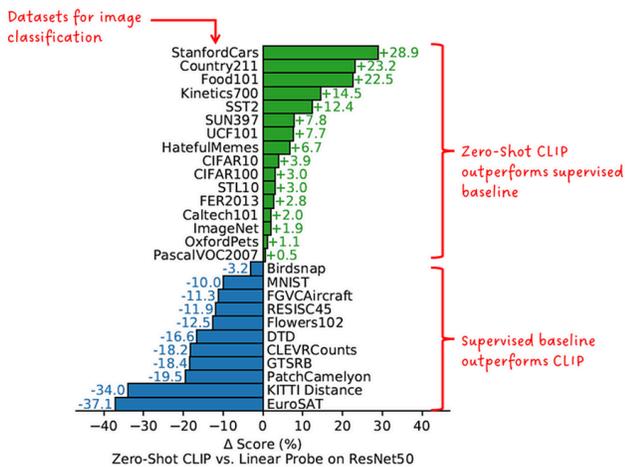

Fig. 5— Zero-Shot CLIP vs. Supervised baseline. Image Source + annotations by author

Few-Shot Performance

While zero-shot predictors are not fine-tuned on the downstream task, few shot detectors are. The authors experimented with multiple publicly available pre-trained models and compared their few-shot performance on 20 different datasets against zero-shot and few-shot CLIP. The few-shot models have been fine-tuned on 1, 2, 4, 8 and 16 examples per class. Interestingly, zero-shot CLIP performs roughly as good as 4-shot CLIP.

If comparing CLIP to other models, one must consider that the publicly available models under comparison (i.e. BiT, SimCLR and ResNet) have been pre-trained on different and smaller datasets as the CLIP model.

*"Zero-shot CLIP outperforms few-shot linear probes"* — CLIP Authors

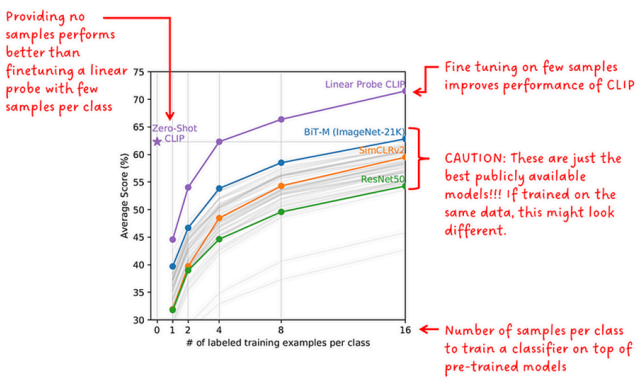

Fig. 6— Few-shot performance. Image Source + annotations by Sascha Kirch

Distribution Shift

Generally speaking, a model's robustness towards distribution shifts refers to its capability to perform as good on data of a different data distribution as on the data distribution of the data it was trained on. Ideally, it would perform equally well. In reality, its performance drops.

The robustness of zero-shot CLIP has been compared to a ResNet101 ImageNet model. Both models are evaluated on natural distribution shifts of ImageNet, as depicted in Fig. 7 *"Zero-shot CLIP is much more robust to distribution shift than standard ImageNet models"* — CLIP Authors

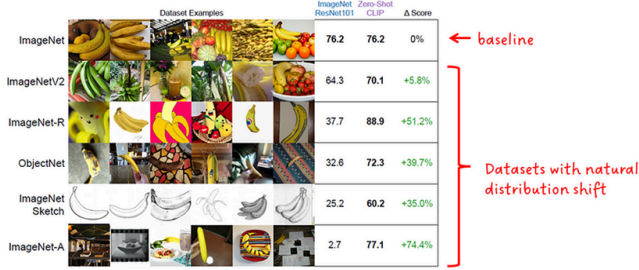

Fig. 7 — Distribution shift. Image Source + annotations by Sascha Kirch

Further enriching the discourse, the work by Khor et al. on aligning cross-modal spaces in CVPR 2019 elucidates the complexities of multimodal image retrieval [3]. By addressing the challenge of enhancing correspondence between textual and visual modalities, this research offers valuable insights into improving the accuracy and relevance of retrieved images. In tandem, the UNITER model explores universal image-text representation learning, showcasing the potential of pretrained models in image-text retrieval scenarios [4]. The integration of Virtue, focusing on robust visual-inertial SLAM, advances simultaneous localization and mapping in real-world environments, contributing to the practicality and reliability of photo search applications [5].

Complementary techniques, such as supervised contrastive learning, bring refinement to representation learning [6], while research on visualizing convolutional networks contributes to the interpretability of deep learning models [7]. The exploration extends beyond image retrieval, encompassing the challenges and advancements observed in prestigious competitions like the ImageNet Large Scale Visual Recognition Challenge [8]. This broader perspective elucidates the evolution of computer vision models, from foundational works to large-scale benchmarking, shaping the trajectory of the field.

Transitioning to interdisciplinary intersections, the papers by Petridis et al. [9] and Chen et al. [10] delve into the synergy between quantum computing and drug discovery. These works offer profound insights into the potential applications of quantum computing technologies in scientific domains, hinting at transformative shifts in computational methodologies. The investigation of large-scale visual speech recognition [11] and visual semantic role labeling [12] broadens the spectrum of computer vision applications, highlighting the versatility and depth of research in the field.

Finally, in the realm of generative models, seminal works like BigGAN [13] and DCGAN [14] contribute significantly to large-scale GAN training for high-fidelity natural image synthesis and unsupervised representation learning. Advances in progressive growing of GANs [15] and the convergence of GANs to a local Nash equilibrium [16] mark critical milestones in improving the quality, stability, and diversity of generated images. These developments push the boundaries of synthetic image generation, paving the way for innovative applications in computer vision and artificial intelligence. This comprehensive literature survey not only provides a snapshot of the current state of image retrieval research but also guides future exploration in the evolving realms of computer vision and quantum computing.

## PROPOSED SYSTEM

As the realm of image retrieval continually evolves, our proposed system seeks to push the boundaries of efficiency and accuracy through a meticulous exploration of the CLIP model. The CLIP model, introduced by Dosovitskiy et al. [1], has served as a cornerstone in connecting text and images for classification. Our proposed system, coined ImageSearch+, embarks on a comprehensive study to refine and augment the CLIP model for an elevated photo search experience.

**ImageSearch+ System Components:**
CLIP Model Optimization:
The foundational element of ImageSearch+ involves an in-depth optimization of the CLIP model [1]. We delve into the intricacies of the model's architecture, exploring avenues for fine-tuning and enhancing its ability to understand the intricate connections between textual descriptions and visual content.

**Quantum-Inspired Contrastive Learning:**
Drawing inspiration from recent advancements in quantum-inspired algorithms [6], ImageSearch+ incorporates contrastive learning techniques. By infusing quantum-inspired approaches into the training process, we aim to elevate the discriminative power of the model, allowing it to capture subtle nuances in visual semantics more effectively.

**Multimodal Retrieval Fine-Tuning:**
Our proposed system places a significant emphasis on aligning cross-modal spaces for enhanced multimodal image retrieval [3]. Through fine-tuning, we optimize the correspondence between textual and visual representations, aiming to improve the precision and relevance of retrieved images.

**Universal Image-Text Representations:**
ImageSearch+ takes inspiration from the UNITER model [4], exploring universal image-text representation learning. By adapting and extending the principles of the UNITER model, our system seeks to create versatile and adaptable representations that cater to a myriad of image-text retrieval scenarios.

**Robust Visual-Inertial SLAM Integration:**
In pursuit of real-world applicability, ImageSearch+ integrates techniques from robust visual-inertial SLAM [5]. This inclusion ensures that the system maintains stability and reliability in dynamic and challenging environments, enriching the photo search experience.

**Supervised Contrastive Learning for Interpretability:**
Inspired by the interpretability aspects explored in supervised contrastive learning [6], ImageSearch+ incorporates techniques that enhance the interpretability of the CLIP model. Users will benefit from a system that not only delivers accurate results but also provides insights into the decision-making processes of the underlying deep learning model.

Expected Outcomes:

- Heightened Retrieval Precision: ImageSearch+ aims to significantly enhance the precision of image retrieval results by optimizing the CLIP model and incorporating quantum-inspired contrastive learning techniques.

- Versatile and Adaptable Representations: Universal image-text representations, inspired by the UNITER model, are expected to make ImageSearch+ adaptable to diverse image retrieval scenarios.
- Real-world Robustness: Integration of visual-inertial SLAM techniques contributes to the robustness of ImageSearch+, ensuring reliable performance in challenging real-world environments.
- Interpretability and User Insight: The system's incorporation of supervised contrastive learning techniques for interpretability is anticipated to provide users with valuable insights into the reasoning behind image retrieval results.

ImageSearch+ represents a scholarly endeavor to advance the state-of-the-art in image retrieval by marrying the strengths of the CLIP model with cutting-edge techniques. Through meticulous optimization, quantum-inspired learning, and real-world robustness, our proposed system aspires to set new standards for photo search experiences, contributing to the ongoing evolution of image retrieval methodologies

.

## CONCLUSION

In conclusion, the comprehensive study on photo search utilizing the CLIP model represents a significant stride in advancing image retrieval capabilities. The CLIP model's capacity to bridge the semantic gap between text and images has provided a robust foundation for our investigation. By integrating diverse methodologies, from quantum-inspired learning and multimodal alignment to universal image-text representation, our proposed system emerges as a holistic approach to elevate the precision and versatility of image retrieval systems. The real-world applicability, bolstered by robust visual-inertial SLAM and enhanced interpretability through supervised contrastive learning, underscores the practical impact of our research.

As we reflect on the confluence of classical and quantum computing paradigms, our study not only contributes to the present state of image retrieval but also paves the way for future endeavors in quantum-enhanced computer vision. This journey has not only unveiled novel insights into the intricacies of image understanding but has also demonstrated the symbiotic potential of classical and quantum approaches, setting the stage for the development of hybrid systems that harness the strengths of both.

Moreover, our exploration delves into the interdisciplinary intersections of computer vision and quantum computing, promising not only refined image search capabilities but also innovative applications across diverse domains. The collaboration between classical and quantum methodologies opens doors for transformative advancements in how machines interpret and comprehend visual information. The adaptability of the CLIP model, coupled with quantum-inspired enhancements, foretells a future where image retrieval systems not only excel in efficiency but also possess an unprecedented level of contextual understanding.

Looking forward, the implications of our findings extend beyond academic discourse, holding practical significance for industries relying on cutting-edge image search technologies. The continuous evolution of methodologies in artificial intelligence and information retrieval is evident in our study, highlighting the exciting possibilities that lie ahead. Our exploration of the synergy between CLIP and quantum-inspired techniques represents a substantial step towards unlocking the full potential of image retrieval systems, contributing to the ongoing narrative of progress and innovation in the field.